%% file: acl_latex.tex
\pdfoutput=1

\documentclass[11pt]{article}

\usepackage[final]{acl}

\usepackage{times}
\usepackage{latexsym}

\usepackage[T1]{fontenc}

\usepackage[utf8]{inputenc}

\usepackage{microtype}
\usepackage{booktabs}
\usepackage{multirow} 
\usepackage{inconsolata}

\usepackage{graphicx}
\usepackage{adjustbox}
\usepackage{svg}
\usepackage[most]{tcolorbox}

\title{Small Language Models in the Real World: Insights from Industrial Text Classification}

\author{
 \textbf{Lujun Li\textsuperscript{1}},
 \textbf{Lama Sleem\textsuperscript{1}},
 \textbf{Niccolo' Gentile\textsuperscript{2}},
 \textbf{Geoffrey Nichil\textsuperscript{2}},
  \textbf{Radu State\textsuperscript{1}}
\\
 \textsuperscript{1}University of Luxembourg,
 \textsuperscript{2}Foyer S.A.,
\\
 \small{
   \textbf{Correspondence:} \href{mailto:lujun.li@uni.lu}{lujun.li@uni.lu}
 }
}

\begin{document}
\maketitle
\begin{abstract}

With the emergence of ChatGPT, Transformer models have significantly advanced text classification and related tasks. Decoder-only models such as Llama exhibit strong performance and flexibility, yet they suffer from inefficiency on inference due to token-by-token generation, and their effectiveness in text classification tasks heavily depends on prompt quality. Moreover, their substantial GPU resource requirements often limit widespread adoption. Thus, the question of whether smaller language models are capable of effectively handling text classification tasks emerges as a topic of significant interest. However, the selection of appropriate models and methodologies remains largely underexplored. In this paper, we conduct a comprehensive evaluation of prompt engineering and supervised fine-tuning methods for transformer-based text classification. Specifically, we focus on practical industrial scenarios, including email classification, legal document categorization, and the classification of extremely long academic texts. We examine the strengths and limitations of smaller models, with particular attention to both their performance and their efficiency in \textbf{V}ideo \textbf{R}andom-\textbf{A}ccess \textbf{M}emory (VRAM) utilization, thereby providing valuable insights for the local deployment and application of compact models in industrial settings\footnote{\scriptsize \url{https://github.com/DobricLilujun/agentCLS/}}\footnote{This paper has been accepted as a conference paper in the Industry Track of the 63rd Annual Meeting of the Association for Computational Linguistics (ACL).}.
\end{abstract}

\section{Introduction}

Text classification is a fundamental task in natural language processing (NLP) that involves the automatic assignment of textual documents, regardless of length, to predefined categories \cite{Taha_2024}. With the exponential growth of digital textual data, the significance of this task has increased considerably. Efficient classification methods have become increasingly valuable in both academic research and industrial applications, while the complexity of classification has also escalated \cite{DBLP:journals/corr/abs-1811-01910}. The field has evolved from basic sentiment analysis of entire texts to more advanced approaches such as multi-label classification and hierarchical classification of long documents\cite{wang-etal-2023-towards-better}. These advancements have led to greater demands for customization and higher classification efficiency, particularly in industrial applications. In scenarios with abundant labeled data, certain encoder-only models can be quickly trained and deployed. However, in cases with limited or no labeled samples, BERT-like models \cite{DBLP:journals/corr/abs-1810-04805} often struggle to achieve satisfactory performance. For localized industrial deployments, achieving optimal results typically requires large-scale models like Llama-3.1-70B-Instruct, which demands significant GPU resources. This makes their widespread use in industrial text classification less practical compared to models like BERT, as dedicating high-memory GPUs solely for classification is often infeasible.



As a consequence, this study aims to investigate the limitations of transformer models, with a particular focus on the performance of Small Language Models (SLMs) and exploring best practices to address industrial text classification challenges effectively. To achieve this, we center our research around three key questions:

\begin{itemize}
    \item \textbf{RQ1}: Can SLMs perform classification without any task-specific training?
    \item \textbf{RQ2}: What are the strengths and limitations of various methods applied to text classification using SLMs?
    \item \textbf{RQ3}: How can the trade-off between computational efficiency and classification performance be optimized, and how can SLMs be more effectively deployed in practice?
\end{itemize}

The remainder of this paper is organized as follows. Section~2 reviews related work and text classification approaches; Section~3 presents the experimental methodology applied to industrial datasets; Section~4 provides a detailed analysis of the results; and Section~5 concludes the study with key findings and future directions.

\section{Related Work}
\input{sections/RW}

\section{Experiments On Industrial Cases}
\input{sections/EOIC}

\section{Results}
\input{sections/RESULTS}

\section{Conclusion}

In this study, we present a comprehensive evaluation of lightweight models on text classification. We systematically investigate nearly all major approaches, including prompt engineering and supervised fine-tuning. Our experimental setup spans three benchmark datasets, including a real-world industrial scenario involving email history classification.

Our findings indicate that while the volume of training data has a significant impact on classification performance, the model's intrinsic understanding of domain-specific textual content also plays a critical role and can become a major bottleneck in achieving high accuracy. Furthermore, we observe that increasing the size of the model or the depth of the classification head yields only marginal performance improvements.

Finally, we analyze the VRAM efficiency of different models across the entire classification pipeline, offering practical insights into their suitability for real-world deployment. These results are particularly relevant for industrial applications, where both high precision and computational efficiency are essential, providing guidance in selecting the appropriate models, classification strategies, and computational resources to optimize under real-world constraints.

\section{Limitations}

This paper comprehensively evaluates Transformer-based classification methods on industrial datasets, providing valuable insights for real-world deployment. However, the impact of the number of virtual tokens in SFT has not been thoroughly explored. It is possible that increasing the number of virtual tokens could yield better results.

Furthermore, we observed that the performance of the ModernBERT-base model on the EUR dataset is particularly poor. However, due to the limited understanding of its pretraining data volume and composition, further research is needed to analyze the language understanding capabilities of ModernBERT-base. Since our training does not enhance the model’s intrinsic language understanding, the model’s inherent linguistic comprehension plays a crucial role in classification tasks. Additionally, more SLMs should be evaluated, such as Gemma-2B, to obtain a more comprehensive understanding of the results.

\section*{Acknowledgments}

This research has benefited from the collaboration and support of our industrial partner, academic institutions, and contributors. We thank the ``AI \& Data Studio'' team for their insights, guidance, and provision of essential computational resources (NVIDIA H100 GPUs), which were crucial for the experiments. The mentorship from faculty members and feedback from postdoctoral researchers have greatly improved the study's rigor. Additionally, we confirm that no AI-generated text was used in preparing this manuscript. Our draft complies with the European General Data Protection Regulation
 (GDPR) data policy.

\bibliography{custom}

\appendix
\section{Experiment Details}
\label{apdx_experimental_details}

In this study, we examine three distinct models in all text classification methods, along with several larger models, as presented in Table \ref{tab:model_specs}.

\begin{table*}[]
\centering
\begin{adjustbox}{max width=\textwidth}
\begin{tabular}{lcccc}
\toprule
\textbf{Model} & \textbf{Ctx Len} & \textbf{Release} & \textbf{VRAM Train(GB)} & \textbf{VRAM Infer(GB)} \\
\midrule
Llama-3.2-1B-Instruct  & 128k  & Sep 25, 2024 & 27.36  & 25.78  \\  
Llama-3.2-3B-Instruct  & 128k  & Sep 25, 2024 & 65.52 & 39.55  \\  
ModernBERT-base        & 8,192 & Dec 19, 2024 & 12.82 & 1.72   \\  
ModernBERT-large       & 8,192 & Dec 19, 2024 & 25.48 & 3.35   \\  
Llama-3.3-70B-Instruct & 128k  & Mar 14, 2025 & N/A    & 168     \\  
GPT4o-mini             & 32k   & Jul 18, 2024 & N/A    & N/A     \\  
\bottomrule
\end{tabular}
\end{adjustbox}
\caption{Table of Model Specifications with GPU Memory Requirements. In this table, ``Ctx'' Len refers to the maximum context length, ``Release'' denotes the model's release date, ``VRAM Train (GB)'' indicates the amount of VRAM required for training each model with a batch size of 8 and a context length of 4096, and ``VRAM Infer (GB)'' specifies the VRAM needed to load the model and perform inference.}
\label{tab:model_specs}
\end{table*}

We primarily utilized the AutoModelForSequenceClassification from Transformers to train our model for classification tasks. The main principle involves adding a linear mapping head for model classification, where the input dimension corresponds to the output dimension of the LLMs. For instance, in the case of Llama-3.2-1B-Instruct, its output features are 2048, which serve as the input features for the linear mapping head. The output features' dimension, on the other hand, corresponds to the number of classification labels.

During training, the orignal weights of the pre-trained model are kept frozen, while only the newly introduced classification head is optimized to achieve the final classification objective. In this study,the optimization process is guided by \textbf{BCEWithLogitsLoss}, which serves as the loss function throughout the training.

\section{Prompt Example}
\label{sec:prompt_example}

The base prompt template for the EUR dataset is shown below. Basically, it requires the models to provide three labels with a classification answer at the end, following a separator \textbf{\#\#\#\#}.

\begin{tcolorbox}
Return the classification answer after a separator \#\#\#\#. Do not return any preamble, explanation, or reasoning.

Classify the \textbf{input} text into one of the following categories based on the descriptions provided, and explicitly provide the output classification at the end. 

Categories:
1. \textbf{Decision} - Choose this category if the text involves making a choice or selecting an option.
2. \textbf{Directive} - Use this category if the text instructs or commands an action.
3. \textbf{Regulation} - Appropriate for texts that stipulate rules or guidelines.

$<<<$START OF INPUT$>>>$

\{input\}

$<<<$END OF INPUT$>>>$
\end{tcolorbox}

In the LDD dataset, there will be 11 labels, each representing the category of an academic subject, while the input will be the document version of academic articles. The base prompt template for the LDD dataset is shown below.
\newpage

\begin{tcolorbox}
Return the classification answer after a separator \#\#\#\#. Do not return any preamble, explanation, or reasoning.

Classify the \textbf{input} text into one of the following categories based on the descriptions provided, and explicitly provide the output classification at the end. 

Categories:

- **cs.AI**: Involves topics related to Artificial Intelligence.
- **cs.CE**: Related to Computational Engineering.
- **cs.CV**: Pertains to Computer Vision.
- **cs.DS**: Concerns Data Structures.
- **cs.IT**: Deals with Information Theory.
- **cs.NE**: Focuses on Neural and Evolutionary Computing.
- **cs.PL**: Involves Programming Languages.
- **cs.SY**: Related to Systems and Control.
- **math.AC**: Pertains to Commutative Algebra.
- **math.GR**: Involves Group Theory.
- **math.ST**: Related to Statistics Theory.

$<<<$START OF INPUT$>>>$

\{input\}

$<<<$END OF INPUT$>>>$
\end{tcolorbox}

In the real-world IE dataset, we used authentic email history records from the industry as the data source, with labels manually identified by experts from our industrial partners.

Particularly of interest, we consider Self-consistency COT method to further validate the model's logical reasoning ability. In this approach, the model first generates three different reasoning chains using a COT prompt. Then, the reasoning chains, along with the question, are presented to the model, which selects the most consistent reasoning chain and ultimately identifies the correct classification label.

\begin{tcolorbox}
Return the classification answer after a separator \#\#\#\#. Do not return any preamble, explanation, or reasoning. 

You will be provided three thinking paths for answering the text classification question, and the conclusions from the three paths will be compared. If two or more paths arrive at the same classification result, that will be selected as the most consistent answer; if all three paths differ, answer with the most plausible classification based on the overall reasoning. The self consistency prompt template is shown below.

Question:

\{question\}

Path 1:
\{path 1\}

Path 2:
\{path 2\}

Path 3:
\{path 3\}

\end{tcolorbox}

\section{Additional Results}

\begin{table*}[!h]
\centering
\adjustbox{valign=c}{
\begin{tabular}{cccccccc}
\toprule
\textbf{\multirow{2}{*}{Methods}} & \textbf{\multirow{2}{*}{Models}} & \multicolumn{2}{c}{\textbf{EUR}} & \multicolumn{2}{c}{\textbf{LDD}} & \multicolumn{2}{c}{\textbf{IE}} \\
\cmidrule(lr){3-8}
\multicolumn{1}{l}{} & \multicolumn{1}{l}{} & \textbf{ACC} & \textbf{F1} & \textbf{ACC} & \textbf{F1} & \textbf{ACC} & \textbf{F1} \\
\midrule
\multicolumn{1}{l}{} & GPT4o-mini & \textbf{0.833} & \textbf{0.767} & 0.682 & 0.698 & \textbf{-} & \textbf{-} \\
\multicolumn{1}{l}{} & Llama-3.3-70B-Instruct & 0.398 & 0.287 & 0.500 & 0.333 & 0.800 & 0.799 \\
\midrule
Base prompt & Llama-3.1-8B-Instruct & 0.216 & 0.193 & 0.554 & 0.596 & 0.500 & 0.333 \\
Few-shot Prompt & Llama-3.1-8B-Instruct & 0.494 & 0.460 & 0.456 & 0.490 & 0.530 & 0.408 \\
Chain-of-Thought & Llama-3.1-8B-Instruct & 0.503 & 0.465 & 0.650 & 0.656 & 0.514 & 0.423 \\
Self-consistency COT & Llama-3.1-8B-Instruct & 0.568 & 0.528 & 0.231 & 0.248 & 0.500 & 0.333 \\
Chain-of-Draft & Llama-3.1-8B-Instruct & 0.422 & 0.375 & \textbf{0.622} & \textbf{0.635} & 0.498 & 0.332 \\
\bottomrule
\end{tabular}
}
\caption{This table presents the performance results of all prompt engineering tests conducted on the larger-scale model, Llama-3.1-8B-Instruct.}
\label{tab:llama}

\end{table*}

We conducted a comprehensive evaluation of various prompt engineering techniques on the relatively large-scale model, Llama-3.1-8B-Instruct, with the aim of achieving competitive performance in comparison to other SLMs. As shown in Table \ref{tab:llama}, despite leveraging an 8-billion parameter model, attaining satisfactory accuracy proved challenging. Notably, the performance improvements achieved through COT and COD strategies were significantly more substantial, markedly outperforming those obtained via Few-shot Prompting. This suggests that for larger models, COT and COD methodologies should be prioritized, whereas few-shot prompting remains the optimal approach for smaller models.

Furthermore, it is important to highlight the poor performance of Self-Consistency COT on the LDD dataset. This limitation is primarily attributed to the excessively long text sequences within LDD, which induce hallucination effects in the model. Given that Self-Consistency COT involves generating three separate reasoning chains, the input length increases considerably, leading to a noticeable degradation in performance. In contrast, COD demonstrates comparable performance to GPT-4o-mini on the LDD dataset, indicating its potential as a promising area for further investigation.

\end{document}

%% file: sections/RW.tex
\subsection{Different Types of Transformers}

Transformers have demonstrated remarkable efficacy in classification tasks \cite{zhao2023transformer}, primarily due to their ability to comprehend multilingual texts and generate linguistically nuanced and stylistically personalized outputs \cite{zhao2024largelanguagemodelshandle}. Across encoder-decoder architectures of LLMs, three primary paradigms emerge: 

1. The sequence to sequence framework \cite{naveed2024comprehensiveoverviewlargelanguage} maps an input sequence to a hidden space, enabling various downstream tasks by appending additional components of the neural network, such as the classifier head. This framework encompasses a range of models, including T5 \cite{DBLP:journals/corr/abs-1910-10683}, and BART \cite{DBLP:journals/corr/abs-1910-13461}, which have been extensively employed in applications such as machine translation and text summarization. 


2. Encoder-only models, such as BERT \cite{devlin2019bertpretrainingdeepbidirectional}, are designed to focus on understanding and processing input text to extract meaningful representations. They demonstrated superior performance in tasks such as named entity recognition (NER: \cite{DBLP:journals/corr/abs-2112-00405}), surpassing other state-of-the-art (SOTA) models. Additionally, models like RoBERTa (Robustly Optimized BERT \cite{DBLP:journals/corr/abs-1907-11692}) and ModernBERT \cite{warner2024smarterbetterfasterlonger} (149M parameters) are optimized for lightweight deployment due to their smaller size.

3. Decoder-only models, with a more compact structure \cite{gao2022encoderdecoderredundantneuralmachine}, extract linguistic knowledge from large corpora and generate translations auto-regressively. They have shown strong performance in text generation \cite{hendy2023goodgptmodelsmachine, brown2020languagemodelsfewshotlearners}. The rapid growth of language models is driven by decoder-only architectures, known for their versatility, reasoning, and problem-solving abilities. Their decoding mechanism allows them to handle nearly all NLP tasks. Notable examples include Meta’s Llama series \cite{touvron2023llamaopenefficientfoundation} and Google’s Gemma series \cite{team2024gemma}, along with newly released reasoning models such as DeepSeek \cite{liu2024deepseek}, which enhance logical problem-solving by leveraging hard-coded reasoning chains.

\subsection{Background}

The earliest systematic studies on text classification included probabilistic model-based methods such as Naive Bayes \cite{SVMTCS}. He was the first to apply Support Vector Machines (SVM) to text classification tasks. With the advent of neural networks, early research primarily utilized embeddings and simple neural network architectures for text classification. Subsequently, \cite{DBLP:journals/corr/Kim14f} proposed a convolutional neural network-based approach for text classification, significantly improving classification performance at sentence-level feature extraction. In addition, classification models based on Recurrent Neural Networks (RNNs) have also shown remarkable performance, demonstrating greater robustness under distribution shifts \cite{yogatama2017generativediscriminativetextclassification}. However, they still struggle to effectively handle complex scenarios in classification tasks such as long texts\cite{du-etal-2020-pointing}. Later, the emergence of attention architectures led to extensive experimentation in various applications.

The advent of transformer-based architectures in 2018, particularly BERT, brought about a paradigm shift in natural language classification tasks, resulting in considerable performance enhancements \cite{10278387, pawar-etal-2024-generate}. Some knowledge distillation approaches \cite{nityasya2022studentbestcomprehensiveknowledge} have also been explored to compress large BERT models into smaller, faster, and more efficient versions that can retain up to 97\% of the original model's classification performance. This observation has motivated our interest in directly using small open source models, which often achieve performance comparable to that of large models after distillation \cite{zhu-etal-2024-survey-model}. For long text classification, specialized bidirectional models such as Longformer \cite{DBLP:journals/corr/abs-2004-05150} and LegalBERT \cite{chalkidis-etal-2020-legal} have emerged in recent years, capable of handling ultra-long documents and showing excellent performance. Nevertheless, their adoption in industry remains limited, primarily due to substantial GPU resource requirements and the need for custom CUDA kernels to support sliding-window attention, which also introduces compatibility challenges with the Huggingface Transformers framework.

Regarding SLMs, \cite{lepagnol-etal-2024-small} explored the zero-shot text classification capabilities of small language models, highlighting their potential in classification tasks. Recent advancements in text classification have primarily focused on two key approaches: prompt engineering and \textbf{S}upervised \textbf{F}ine \textbf{T}uning(SFT).

Prompt engineering involves crafting well-structured inputs to guide LLMs in producing more personalized responses. Recent research has shown that sophisticated prompt engineering techniques can sometimes compete with or even outperform fine-tuned models\cite{sahoo2025systematicsurveypromptengineering}. In both industry and academia, models such as BERT and Llama are commonly used to assess downstream tasks. Nevertheless, there is a notable absence of extensive comparative research on various prompt engineering and SFT techniques for SLMs, aimed at identifying the most effective practices for industrial applications. Furthermore, publicly available datasets are frequently subject to inherent biases resulting from prior exposure during pre-training, which means that models being evaluated may have already been trained on portions of the test set, thereby introducing the possibility of biases.

%% file: sections/EOIC.tex
\subsection{Methods}

To address the challenges outlined in the related work, we trained models on datasets of varying difficulty levels, including a proprietary, real-world industrial dataset. Regarding model selection, we primarily focused on decoder-only architectures while incorporating a subset of encoder-only models for validation. In addition, we explore various prompt engineering techniques and examine the impact of different prompt tuning methods, focusing on classification task.  

Table \ref{tab:methods_comparison} presents an overview of different templates and prompt strategies, where all prompts are designed to enforce a structured output format. The base prompt closely resembles a direct label mapping approach, where the model outputs the label it deems most appropriate. Few-shot prompts extend this by incorporating examples alongside descriptions. Furthermore, Chain-of-Thought (COT) and Chain-of-Draft (COD) prompts serve to evaluate the reasoning capabilities of SLMs to some extent.

In the training process, we primarily employ three distinct methods: 1) SFT, which modifies only the weights of the classification heads added at the end of the model using labeled data; 2) Soft Prompt Tuning (SPT), which involves optimizing input prompts to continuously guide the model towards correct behavior based on labeled data; and 3) Prefix Tuning (PT), which incorporates a learnable prefix tensor into each attention layer. 

These approaches enhance the model's classification performance while keeping most of the model weights frozen, which are widely used in industrial use cases.




\begin{table}[!htbp]
\begin{adjustbox}{max width=0.5\textwidth} 
\begin{tabular}{ccccl}
\toprule
Methods Types & Methods & \multicolumn{1}{c}{Reference} \\ \midrule
Prompt Engineering & Base Prompts                    & \cite{ye2024promptengineeringpromptengineer} \\ \midrule
Prompt Engineering & Few-Shot Prompts                & \cite{DBLP:journals/corr/abs-2005-14165} \\ \midrule
Prompt Engineering & Chain-of-Thought (COT)               & \cite{DBLP:journals/corr/abs-2201-11903} \\ \midrule
Prompt Engineering & Self-consistency COT            & \cite{wang2023selfconsistencyimproveschainthought} \\ \midrule
Prompt Engineering & Chain-of-Draft (COD)                 & \cite{xu2025chaindraftthinkingfaster} \\ \midrule
Fine Tuning        & Supervised Fine-tuning          & \cite{parthasarathy2024ultimateguidefinetuningllms} \\ \midrule
Soft Prompt Tuning & Parameter Efficient Fine-tuning & \cite{DBLP:journals/corr/abs-2104-08691} \\ \midrule
Prefix Tuning      & Parameter Efficient Fine-tuning & \cite{DBLP:journals/corr/abs-2101-00190} \\ \bottomrule
\end{tabular}
\end{adjustbox}
\caption{Classification methods based on the transformer architecture investigated in this study.}
\label{tab:methods_comparison}
\end{table}

\subsection{Datasets}

\begin{table*}[htbp]
    \centering
    \adjustbox{max width=1.0\textwidth}{
        \begin{tabular}{ccccccc}
            \toprule
            \textbf{Dataset} & \textbf{Abbreviation} & \textbf{Words / D} & \textbf{\# Train} & \textbf{\# Validation} & \textbf{\# Labels} & \textbf{Subject} \\
            \midrule
            EURLEX57K             & EUR & 720   & 3039  & 900  & 3  & EU Legislation \\
            Long Document Dataset & LDD & 10378 & 15682 & 3300 & 11 & Academy        \\
            Insurance Email       & IE  & 724   & 2015  & 1000 & 2  & Email History  \\
            \bottomrule
        \end{tabular}
    }
    \caption{The table below presents the statistics of the three datasets used in our experiments. \texttt{Words/D}  denotes the average number of words per document, \texttt{\#Train}  represents the number of training samples, \texttt{\#Validation}  refers to the number of validation samples, and \texttt{\#Labels} indicates the number of unique labels in the dataset. Each dataset corresponds to a different domain of text. Notably, the LDD dataset exhibits a larger number of labels and a higher word count per document, which increases the difficulty of the classification task.}
    \label{tab:dataset_stats}
\end{table*}

In this study, we primarily utilized three datasets for our experiments, as shown in Table \ref{tab:dataset_stats}. First, we used the EURLEX57K dataset \cite{chalkidis-etal-2019-large}, which was released by \cite{chalkidis-etal-2019-large} and contains 57,000 new legislative documents. We adopted the document type as the classification label, which includes Regulation, Decision, and Directive. Additionally, we employed the Long Document Dataset \cite{8675939}, a relatively more challenging dataset that consists of a large amount of literature text extracted from PDFs, categorized into 11 different classes, such as cs.AI (Artificial Intelligence), cs.CE (Computational Engineering), and so on. The main difficulty lies in the length of the documents and the challenge of classifying them into over 11 labels, which significantly increases the complexity of the task.

In addition, we possess a proprietary, closed-source dataset derived from email correspondence between our partner company and its clients. The primary business requirement is to analyze historical interactions with each client—written in a mixture of English, French, German, and Luxembourgish—to determine whether the most recent emails in the thread are reminders. Consequently, the task involves identifying the optimal position within the text and determining whether that position conveys a ``reminder'' meaning, resulting in a binary labeling scheme. It also requires a comprehensive understanding of long email threads written in mixed languages, including low-resource ones, and making a final decision based on the contextual meaning at the identified position.

The main challenges associated with this dataset are: 1. Semantic decision-making is heavily based on the content of the most recent emails exchanged with the client, with older emails primarily serving as background context. This characteristic places the most crucial textual information towards the beginning of the sequence, which contrasts with typical datasets where classification decisions are based on the overall semantics of the entire text. 2. The dataset inherently contains long texts with uneven length distributions with information extracted from images. All nontextual data has been processed using OCR to extract textual content. By incorporating this real-world industrial dataset, we improve the persuasiveness and robustness of our model and methods evaluations.

\subsection{SLM Models}

Fine-tuning on classification typically refers to the application of transfer learning when a task is associated with a certain amount of labeled data. This approach capitalizes on the semantic representation capabilities of a pre-trained model by incorporating a lightweight linear layer for classification, denoted as classification heads. During training, the model parameters are kept frozen, while only the newly introduced classification network is optimized to achieve the classification objective. In this study, we adopt SLMs including \textbf{Llama-3.2-1B}, \textbf{Llama-3.2-1B} and \textbf{ModernBERT-base} as the foundational models. Additionally, Llama-3.3-70B-Instruct and GPT-4o mini are used as foundation model baselines for performance comparison. More details are shown in the Appendix \ref{apdx_experimental_details}.

\subsection{Experimental Settings \& Metrics}

We employ \textbf{Accuracy}, \textbf{F1-score} as performance metrics to evaluate different methods across all models. For the \textbf{fine-tuning} approach, we standardize the learning rate to \textbf{1e-6} and train all models for \textbf{10 epochs} to ensure controlled variable conditions. To evaluate the efficiency of different methods and analyze resource usage, we track GPU hours (GHs) and GPU RAM hours (GRHs). GPU hours represent the total computational time a model utilizes GPU clusters, while GPU RAM hours quantify cumulative memory consumption during execution. These metrics provide insights into computational cost and resource efficiency. As prompt engineering primarily affects inference time and pretraining duration is unknown, we measure only its inference stage.

The prompts used from different strategy methods were well designed as shown in the appendix \ref{sec:prompt_example}. When it comes to self-consistency COT, several different paths of thinking should be set, and in this study, we explicitly set it to 3. To control for variables, we standardize the batch size to 8 and set the number of training epochs to 10, selecting the checkpoint with the lowest evaluation loss. For both SPT and PT, we configure the number of virtual tokens to 128. In general, all models are trained with a maximum context length of 4096 tokens.






%% file: sections/RESULTS.tex
\subsection{Main Performance}

\begin{table*}[htbp]
\captionsetup{justification=raggedright,singlelinecheck=false}
    \caption{The main results include validation performance on three datasets under different prompt engineering and SFT conditions. ACC represents accuracy, GH indicates GPU hours, and GRH refers to GPU RAM hours for memory usage. Prefix-tuning is unsupported on ModernBERT-base due to model structure incompatibility.}
    \label{tab:table dataset selection}
    \centering
    \begin{adjustbox}{max width=1.0\textwidth}
    \begin{tabular}{ccccccccccccccc}

    \toprule
\textbf{\multirow{2}{*}{Methods Type}} &
\textbf{\multirow{2}{*}{Methods}} &
\textbf{\multirow{2}{*}{Models}} &
\multicolumn{4}{c}{\textbf{EUR}} &
\multicolumn{4}{c}{\textbf{LDD}} &
\multicolumn{4}{c}{\textbf{IE}} \\
\cmidrule(lr){4-15}
\multicolumn{1}{l}{} &
\multicolumn{1}{l}{} &
\multicolumn{1}{l}{} &
\textbf{ACC ↑} &
\textbf{F1 ↑} &
\textbf{GH ↓} &
\textbf{GRH ↓} &
\textbf{ACC ↑} &
\textbf{F1 ↑} &
\textbf{GH ↓} &
\textbf{GRH ↓} &
\textbf{ACC ↑} &
\textbf{F1 ↑} &
\textbf{GH ↓} &
\textbf{GRH ↓} \\
\cmidrule(lr){3-15}
\multicolumn{1}{l}{} &
\multicolumn{1}{l}{} &
GPT-4o-mini &
\textbf{0.833} &
\textbf{0.767} &
N/A &
N/A &
0.682 &
0.698 &
N/A &
N/A &
N/A &
N/A &
N/A &
N/A \\
\multicolumn{1}{l}{} &
   &
Llama-3.3-70B-Instruct &
0.398 &
0.287 &
0.157 &
26.443 &
0.500 &
0.333 &
0.188 &
31.651 &
\textbf{0.800} &
\textbf{0.799} &
0.517 &
86.772 \\

  \midrule
 &
   &
  Llama-3.2-1B-Instruct &
  0.330 &
  0.319 &
  0.010 &
  0.263 &
  0.186 &
  0.159 &
  0.775 &
  19.981 &
  0.500 &
  0.370 &
  0.040 &
  1.034 \\
 &
  \multirow{-2}{*}{Base prompt} &
  Llama-3.2-3B-Instruct &
  0.346 &
  0.220 &
  0.030 &
  1.167 &
  0.314 &
  0.301 &
  0.313 &
  12.385 &
  0.500 &
  0.333 &
  0.047 &
  1.847 \\
 &
   &
  Llama-3.2-1B-Instruct &
  0.387 &
  0.377 &
  0.022 &
  0.578 &
  0.132 &
  0.113 &
  0.574 &
  14.804 &
  0.488 &
  0.338 &
  0.038 &
  0.972 \\
 &
  \multirow{-2}{*}{Few-shot Prompt} &
  Llama-3.2-3B-Instruct &
  0.506 &
  0.499 &
  0.024 &
  0.931 &
  0.471 &
  0.491 &
  0.136 &
  5.376 &
  0.500 &
  0.333 &
  0.044 &
  1.756 \\
 &
   &
  Llama-3.2-1B-Instruct &
  0.463 &
  0.438 &
  0.181 &
  4.659 &
  0.181 &
  0.167 &
  1.248 &
  32.171 &
  0.501 &
  0.339 &
  0.189 &
  4.873 \\
 &
  \multirow{-2}{*}{Chain-of-Thought} &
  Llama-3.2-3B-Instruct &
  0.341 &
  0.293 &
  0.427 &
  16.906 &
  0.365 &
  0.334 &
  0.722 &
  28.544 &
  0.491 &
  0.401 &
  0.519 &
  20.538 \\
 &
   &
  Llama-3.2-1B-Instruct &
  0.433 &
  0.411 &
  0.582 &
  14.997 &
  0.178 &
  0.168 &
  4.231 &
  109.086 &
  0.500 &
  0.333 &
  0.597 &
  15.392 \\
 &
  \multirow{-2}{*}{Self-consistency COT} &
  Llama-3.2-3B-Instruct &
  0.419 &
  0.338 &
  0.982 &
  38.836 &
  0.167 &
  0.168 &
  2.321 &
  91.821 &
  0.510 &
  0.333 &
  0.991 &
  39.192 \\
 &
   &
  Llama-3.2-1B-Instruct &
  0.408 &
  0.395 &
  0.061 &
  1.560 &
  0.226 &
  0.226 &
  0.376 &
  9.702 &
  0.499 &
  0.336 &
  0.105 &
  2.705 \\
\multirow{-10}{*}{\textbf{Prompt Engineering}} &
  \multirow{-2}{*}{Chain-of-Draft} &
  Llama-3.2-3B-Instruct &
  0.351 &
  0.332 &
  0.055 &
  2.191 &
  0.425 &
  0.437 &
  0.390 &
  15.431 &
  0.499 &
  0.335 &
  0.113 &
  4.458 \\
  \midrule
 &
   &
  Llama-3.2-1B-Instruct &
  0.643 &
  0.533 &
  0.848 &
  22.977 &
  0.442 &
  0.429 &
  4.589 &
  124.827 &
  0.506 &
  0.381 &
  0.594 &
  15.914 \\

 &
   &
  Llama-3.2-3B-Instruct &
  0.641 &
  0.524 &
  2.926 &
  169.812 &
 0.136 &
 0.135 &
 8.303 &
 481.701 &
 0.526 &
 0.475 &
 1.396 &
76.384 \\
 &
  \multirow{-3}{*}{Soft Prompt Tuning (SPT)} &
  ModernBERT-base &
  0.332 &
  0.171 &
  0.533 &
  11.903 &
  0.207 &
  0.184 &
  1.374 &
  26.394 &
  0.500 &
  0.333 &
  0.566 &
  12.667 \\
 &
   &
  Llama-3.2-1B-Instruct &
  0.330 &
  0.266 &
  1.580 &
  42.947 &
  0.112 &
  0.107 &
  7.826 &
  212.826 &
  0.502 &
  0.371 &
  0.463 &
  12.530 \\
 &
   &
  Llama-3.2-3B-Instruct &
  0.320 &
  0.300 &
  1.360 &
  83.864 &
  0.128 &
  0.117 &
  16.532 &
  1040.624 &
  0.588 &
  0.536 &
  2.999 &
  172.257 \\
 &
  \multirow{-3}{*}{Prefix Tuning (PT)} &
  ModernBERT-base &
  N/A &
  N/A &
  N/A &
  N/A &
  N/A &
  N/A &
  N/A &
  N/A &
  N/A &
  N/A &
  N/A &
  N/A \\
 &
   &
  Llama-3.2-1B-Instruct &
  \textbf{0.999} &
  \textbf{0.999} &
  0.508 &
  13.813 &
  \textbf{0.892} &
  \textbf{0.890} &
  1.698 &
  40.631 &
  \textbf{0.865} &
  \textbf{0.863} &
  1.008 &
  27.474 \\
 &
   &
  Llama-3.2-3B-Instruct &
  \textbf{0.998} &
  \textbf{0.998} &
  1.750 &
  92.118 &
  \textbf{0.904} &
  \textbf{0.903} &
  3.764 &
  226.869 &
  \textbf{0.960} &
  \textbf{0.960} &
  1.949 &
  123.109 \\
\multirow{-9}{*}{\textbf{Supervised Fine-Tuning}} &
  \multirow{-3}{*}{Fine-tuning (FT)} &
  ModernBERT-base &
  0.333 &
  0.167 &
  0.132 &
  1.849 &
  0.810 &
  0.811 &
  1.762 &
  24.018 &
  0.514 &
  0.408 &
  0.104 &
  1.476 \\
  \bottomrule
\end{tabular}
    \end{adjustbox}
\end{table*}

Additional models were used to validate the test set in order to provide a reference performance for State-of-the-Art (SOTA) models. However, ChatGPT was not evaluated on the IE dataset due to potential data leakage concerns. In contrast, Llama-3.3-70B-Instruct was run locally, allowing for GPU resource estimation and comprehensive metric evaluation. As presented in Table \ref{tab:table dataset selection}, the highest prompt engineering performance was achieved by ChatGPT-o1 mini. Meanwhile, in the IE dataset, which serves as our industrial database, an accuracy score of 0.800 was achieved by Llama-3.3-70B-Instruct. Regarding SLMs, we found the results particularly intriguing, especially in the context of prompt engineering. Given the relatively small size of these models, we did not expect them to achieve high performance. The final results for the 1B and 3B models aligned with our expectations, performing roughly at the level of random guessing. Interestingly, both the 3B and even the 1B models demonstrated a strong preference for few-shot prompting. This approach led to an improvement of over 10\% compared to the base prompt on the EUR and LDD datasets, highlighting the importance of few-shot learning in the application of SLMs, as also emphasized in \cite{brown2020languagemodelsfewshotlearners}. Furthermore, we observed that both COD and COT provided limited improvements. In fact, on the LDD dataset, COD performed worse than COT and was nearly on par with the base prompt. Therefore, the use of COD and COT is not recommended as a solution for classification tasks in SLMs.

In the context of SFT, we observed that SPT outperformed prefix tuning by a significant margin, although it also required substantially more training time. Prefix tuning introduces a trainable part at every layer within the model, whereas SPT only incorporates a soft prompt at the input level. It is possible that SPT better preserves the original language understanding of the model, as it does not alter the overall architecture. In contrast, prefix-tuning's modifications to the attention structure may disrupt the model's inherent linguistic comprehension. Additionally, supervised fine-tuning, which adds a classification head to the end of the model, demonstrated the highest overall performance. Notably, ModernBERT achieved a performance of approximately 0.810 of accuracy on the LDD dataset while requiring less training time and GPU memory, making it a promising candidate for academic English text classification. Limited exposure to French, other multilingual languages, and domain-specific corpora during training \cite{warner2024smarterbetterfasterlonger} led to weaker performance on the IE dataset (primarily in French) and EUR (a domain-specific corpus).

\subsection{Exploratory Results}

\subsubsection{Does data matter?}

\begin{figure*}[htbp]
    \centering
    \includegraphics[width=1.0\textwidth]{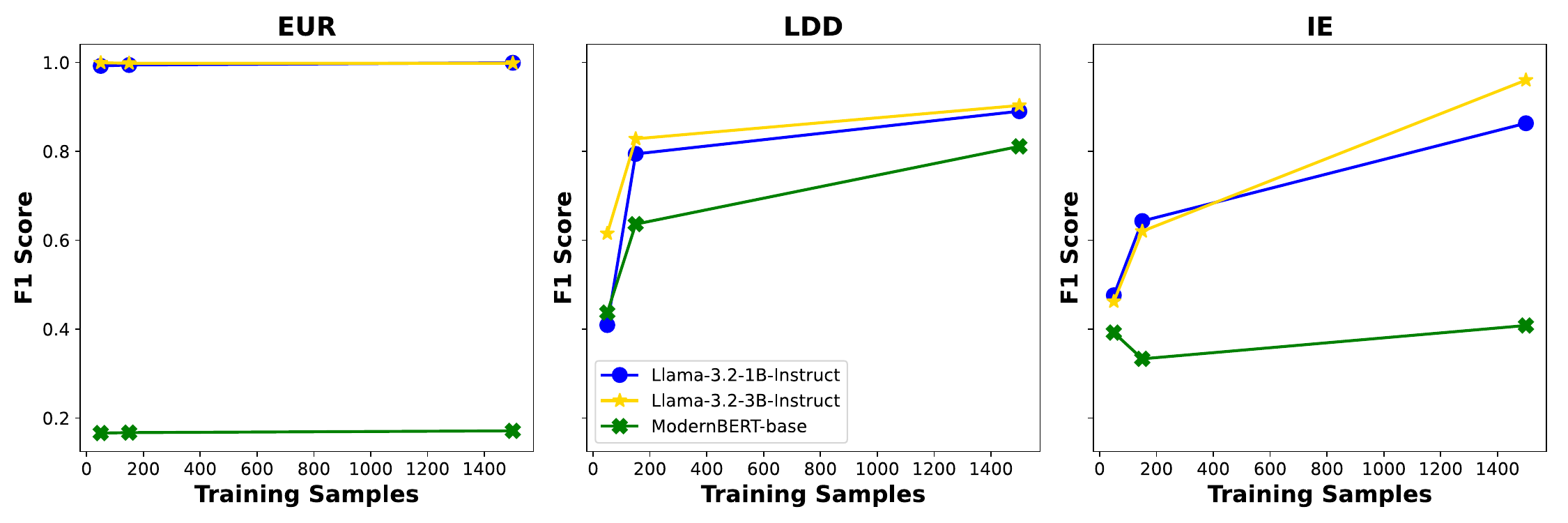} 
    \caption{Impact of Data Volume on Model Performance.}
    \label{fig: data efficiency}
\end{figure*}

Experiments were conducted to examine the impact of data volume, primarily using SFT, the best method in our research scope. We randomly selected 50, 150, and 1500 samples as training data. The results, as shown in Figure \ref{fig: data efficiency}, indicate that on the relatively simple EU dataset, the model can achieve good performance even with a small amount of data after multiple training iterations, with the primary determinant of performance being the model itself. However, for more complex and challenging datasets such as LDD and IE, the amount of training data directly determines performance. Furthermore, we observed that models of different sizes exhibit only minor differences in classification performance. Therefore, data volume has a direct impact on classification performance in difficult datasets, which ultimately defines the performance bottleneck instead of the model itself.




\subsubsection{Larger Models?}

\begin{figure}[!htbp]
    \hspace*{-0.7cm}
    \includegraphics[width=1.2\columnwidth]{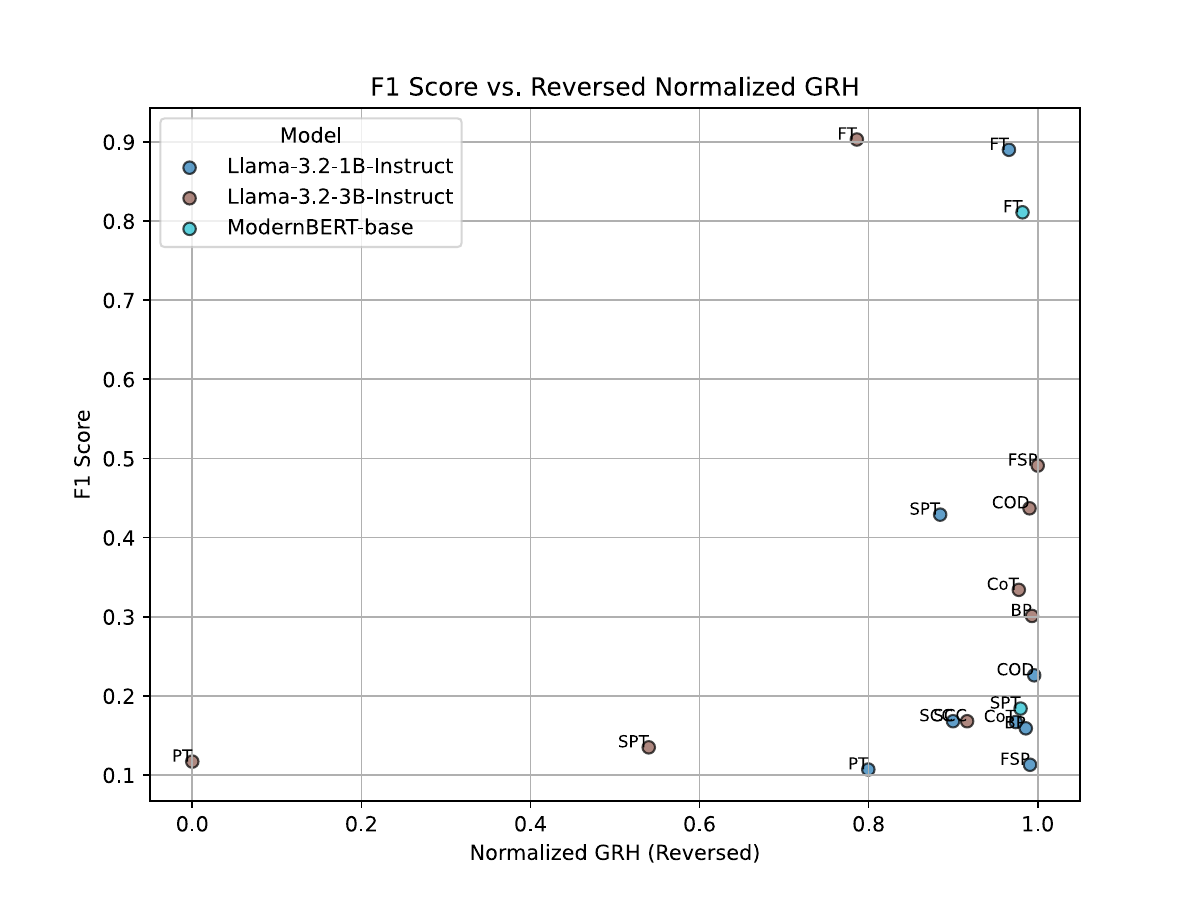} 
    \caption{Reversed efficiency on LDD datasets}
    \label{fig: GRH efficiency on LDD datasets}
\end{figure}

\begin{table}[h]
\centering
\caption{This table compares the performance of ModernBERT-Base ("Base") and ModernBERT-Large ("Large") on the same dataset.}
\label{tab:large or small}
\begin{adjustbox}{max width=0.47\textwidth}
\begin{tabular}{lcc  cc  cc}
\toprule
\textbf{Models} &
  \multicolumn{2}{c}{\textbf{EUR}} &
  \multicolumn{2}{c}{\textbf{LDD}} &
  \multicolumn{2}{c}{\textbf{IE}} \\
\midrule
 & ACC   & F1    & ACC   & F1    & ACC   & F1    \\
\midrule
Base      & 0.333 & 0.167 & 0.810 & 0.811 & 0.514 & 0.408 \\
Large     & 0.333 & 0.168 & 0.828 & 0.829 & 0.539 & 0.424 \\
\bottomrule
\end{tabular}
\end{adjustbox}
\end{table}

As observed in Table \ref{tab:large or small}, the performance gains from larger models are also minimal. For example, in the LDD dataset, ModernBERT-large only improves by about 2\% over the base model. In particular, on the EUR, larger models do not show significant performance gains. This is highly related to the domain relevance of the model’s pre-training data. For example, in the ModernBERT paper, it is mentioned that the model is trained on a large amount of academic English data, which leads to high performance on LDD. The IE dataset, which includes French, German, and English, results in accuracy around 0.5. In the EUR dataset, performance is especially poor and increasing the model size does not improve results. This shows that SFT models for classification do not enhance semantic understanding, but guide comprehension and classification. Thus, the model should be thoroughly investigated before industrial deployment, and decoder-only SLMs are sufficient for classification tasks if they excel at understanding the dataset's domain knowledge. 

\subsubsection{Deeper Header?}

In our primary experimental setting, we adhere to the definition of a ``Header'' as implemented in the Transformers library, referring to a single linear layer serving as the classification head. To further explore potential improvements using different levels of header, we experimented with replacing the standard single-layer header with a multi-layer linear architecture incorporating ReLU activations. Specifically, we constructed classification heads with 2 to 5 linear layers (hidden dimension = 256) and fine-tuned Llama-3.2-1B-Instruct model accordingly. As shown in Table \ref{tab:header_depth_performance}, the results indicate that increasing the depth of the classification head yields only marginal gains, with performance plateauing beyond three layers. These findings suggest that deeper header architectures offer limited benefit in enhancing the classification accuracy or F1 score in this context.

\begin{table}[h]
\centering
\begin{adjustbox}{width=0.8\linewidth}
\begin{tabular}{cccccc}
\toprule
\textbf{\# Layers} & \textbf{1} & \textbf{2} & \textbf{3} & \textbf{4} & \textbf{5} \\
\midrule
\textbf{ACC}       & 0.89       & 0.91       & 0.92       & 0.91       & 0.91       \\
\textbf{F1}        & 0.89       & 0.91       & 0.92       & 0.91       & 0.91       \\
\bottomrule
\end{tabular}
\end{adjustbox}
\caption{Impact of classification head depth on performance, evaluated on the LDD dataset using Llama-3.2-1B-Instruct. ``\# Layers'' refers to the number of stacked linear layers in the classification head.}
\label{tab:header_depth_performance}
\end{table}

\subsection{Efficiency}

We particularly focus on model efficiency from training to inference, with a specific emphasis on VRAM usage, which is the primary limiting factor for deployment in industrial settings. As shown in Figure \ref{fig: GRH efficiency on LDD datasets}, the x-axis represents the reverse normalized GRH score, while the y-axis represents the F1 Score. Therefore, points located further towards the top-right indicate higher efficiency. It is clear that the three FT models exhibit the highest efficiency, while the prompt engineering methods, although very efficient in terms of GPU RAM usage, significantly lag behind in performance. Therefore, for local deployment, fine-tuning of SLMs is the optimal approach for enhancing both efficiency and accuracy. Additionally, we can observe that from 1B to 3B models, there is only a marginal improvement in model accuracy, while GPU time consumption increases. Hence, fine-tuning the 1B model could be the optimal solution when considering efficiency.

\subsection{Research Questions}

For RQ1, ``\textit{Can SLMs perform classification without any task-specific training?}'', we found that text classification using SLMs faces several key challenges. Smaller models tend to exhibit limited logical reasoning capabilities and are more susceptible to generating hallucinations while encountering long text. Moreover, the performance ceiling is strongly influenced by the amount of available training data, while the intrinsic properties of the SLMs themselves also play a critical role in shaping classification outcomes.

Regarding RQ2, ``\textit{What are the strengths and limitations of various methods applied to text classification using SLMs?}'', prompt engineering can demonstrate substantial flexibility and customization; however, its performance on SLMs remains significantly limited. Notably, various prompt engineering strategies, such as COT or COD, sometimes negatively influence model performance. If employing prompts engineering on SLMs is necessary, it is recommended to utilize few-shot prompting rather than COT or COD as shown in Table \ref{tab:table dataset selection}. In contrast, SFT shows excellent performance on decoder-only models, whereas SPT and PT achieve moderate effectiveness. Nevertheless, both approaches generally yield superior results compared to prompt engineering.

For RQ3, ``\textit{How can the trade-off between computational efficiency and classification performance be optimized, and how can SLMs be more effectively deployed in practice?}'', we found that although training the model consumes significant GPU resources, the SLMs are essentially unusable in their current form due to the lack of inference capability. We also tested Llama-3.3-70B-Instruct, which, although capable of achieving 80\% accuracy in IE, still produces uncertain output. Therefore, FT transformers remains the only viable solution on SLMs which is portable and light weight. Finally, the limited capacity of SLMs creates a bottleneck on performance and the amount of labeled data also remains a key limitation. For real application, it is crucial to focus not only on data quality but also on the model's inherent characteristics, such as multilingual comprehension. If resources are relatively abundant, opting for decoder-only models such as the Llama series would be a better choice, which has a good support on both languages and different domain knowledge.